\documentclass{mlcv-fabian}

\begin{document}

\title{\Large\bf Graph Neural Networks with Triangle-Based Messages for
the Multicut Problem}
\author[1]{Jannik Irmai}
\author[1]{Lucas Fabian Naumann}
\author[1,2]{Bjoern Andres}
\affil[1]{Faculty of Computer Science, TU Dresden}
\affil[2]{Center for Scalable Data Analytics and AI, Dresden/Leipzig}
\affil[ ]{\textit{jannik.irmai@tu-dresden.de} \quad \textit{fabian.naumann@tu-dresden.de} \quad \textit{bjoern.andres@tu-dresden.de}}
\date{}
\maketitle

\begin{abstract}
  \noindent The multicut problem is an \textsc{np}-hard combinatorial optimization problem with diverse applications in fields such as bioinformatics, data mining and computer vision.
  Graph neural networks have been defined for the multicut
  problem but can be adapted further to its specific objective function
  and constraints.
  In this article, we introduce such an adapted graph neural network architecture in which features are assigned only to edges, and the computation of messages is based on triangles in the underlying graph.
  Experiments with synthetic and real-world instances with up to 200 nodes show that our method outperforms state-of-the-art heuristic solvers in terms of solution quality while maintaining feasible runtimes.
  For some instances, our method finds optimal solutions in seconds whereas exact solvers need hours to find and certify optimal solutions.
\end{abstract}

\section{Introduction}

The \emph{multicut problem} \citep{chopra1993partition} is a combinatorial optimization problem whose feasible solutions relate one-to-one to the clusterings of a graph.
In particular, for any clustering, the corresponding \emph{multicut} is the set of all edges that straddle distinct clusters.
These edges are said to be \emph{cut} by the multicut, the remaining edges are said to be \emph{joined}.
Given a graph with costs (real numbers) assigned to the edges, the goal of the multicut problem is to find a multicut such that the cost of the cut edges is minimized.
The problem is equivalent to the correlation clustering problem \citep{bansal2004correlation} and the clique partitioning problem \citep{groetschel1990facets}, in the sense that they share optimal solutions. 
In prominent difference to other clustering formulations, the number of clusters is not fixed in advance, but inferred from the data.

The multicut problem and its extensions are used in various fields such as bioinformatics \citep{wolny2020accurate, vergara2021whole}, data mining \citep{shi2021scalable, kostyukhin2023improving} and computer vision \citep{tang2017multiple, nguyen2022lifted}.
Although the multicut problem is \textsc{np}-hard, there exist exact solvers that achieve feasible runtimes for relevant instances \citep{kappes2011globallyoptimal, andres2012globally, letchford2024separation, irmai2025cutting}.
For other instances where these are no longer feasible, a variety of heuristic solvers have been proposed that produce high-quality solutions empirically, but have no approximation guarantees \citep{beier2014cut,  beier2015fusion, keuper2015efficient, levinkov2017comparative, wolf2018mutex, abbas2022rapid, abbas2023clusterfug}.

Since recently, \emph{graph neural networks} (GNNs) \citep{scarselli2009graph} are used to solve combinatorial optimization problems, either directly \citep{selsam2019learning,prates2019learning,toenshoff2021graph}, or by guiding existing solvers \citep{Gasse2019Exact,labassi2022learning}.
There are two approaches that focus specifically on heuristically solving the multicut problem \citep{jung2023learning, li2025dgrl}.
Both use generic GNNs that primarily operate on node features, while the multicut problem is fundamentally edge-based.

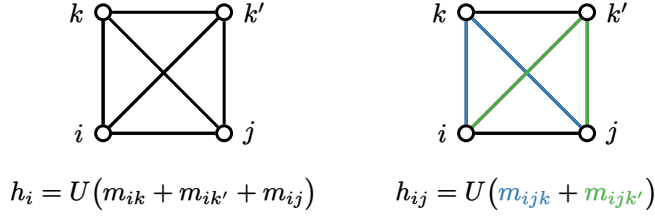
\begin{figure}[t]
    \centering
    \begin{tikzpicture}[scale=1.6, style=base]
    \def \r {0.2cm}

    \node[style=vertex, label=left:$i$] (0) at (3, 0) {};
    \node[style=vertex, label=left:$k$] (1) at (3, 1) {};
    \node[style=vertex, label=right:$j$] (2) at (4, 0) {};
    \node[style=vertex, label=right:$k'$] (3) at (4, 1) {};

    \draw (0) -- (1) -- (2) -- (3) -- (0);
    \draw (0) -- (2);
    \draw (1) -- (3);
    
    \draw[myblue] (0) -- (1) -- (2);
    \draw[mygreen] (0) -- (3) -- (2);
    \draw[] (0) -- (2);

    \node(t) at (3.5, -0.5) {${h_{ij}}=U\bigl({\color{myblue}m_{ijk}} + {\color{mygreen}m_{ijk'}}\bigr)$};

    \node[style=vertex, label=left:$i$] (0) at (0, 0) {};
    \node[style=vertex, label=left:$k$] (1) at (0, 1) {};
    \node[style=vertex, label=right:$j$] (2) at (1, 0) {};
    \node[style=vertex, label=right:$k'$] (3) at (1, 1) {};

    \draw (0) -- (1) -- (2) -- (3) -- (0);
    \draw (0) -- (2);
    \draw (1) -- (3);
    
    \draw[] (0) -- (1);
    \draw[] (0) -- (3);
    \draw[] (0) -- (2);

    \node(t) at (0.5, -0.5) {$h_{i}=U\bigl({m_{ik}} + {m_{ik'}} + {m_{ij}}\bigr)$};
\end{tikzpicture}
    \caption{
        Depicted on the left is an example of the standard message passing scheme. 
        Each node $i$ receives messages from its neighbors, which are aggregated to node features $h_i$.
        Depicted on the right is an example of the triangle message passing scheme we propose.
        Each edge $ij$, receives messages from all triangles containing it (blue and green), which are aggregated to edge features $h_{ij}$.
        }
    \label{fig:messages}
\end{figure}

In this article, we propose a GNN-based heuristic solver, where the GNN architecture is specifically adapted to the multicut problem.
In particular, the architecture is based on the fact that each instance of the multicut problem w.r.t. a graph can be transformed to an equivalent instance on a complete graph \citep{chopra1993partition}, for which the constraints that define feasible solutions are based exclusively on triangles.
Utilizing this, the main building blocks of our GNN architecture are \emph{triangle message passing layers} which operate only on edge features and compute messages based on triangles (see \Cref{fig:messages}).
Thus, they directly capture the constraints of the problem.

We use supervised learning to train our GNN to predict logits for edges being joined in an optimal solution.
For inference, we construct a solution to the multicut problem by alternately applying our model and contracting the edge with the highest logit until all logits become non-positive.
The edges in the remaining graph then correspond to a multicut of the original graph, which is returned as a solution.
We evaluate our approach on random instances and instances from the CP-Lib benchmark \citep{sorensen2024cplib} with up to $200$ nodes and demonstrate that it outperforms state-of-the-art heuristic solvers in terms of solution quality while maintaining feasible runtimes.
Furthermore, we show that our model can find optimal solutions for specific instances in seconds, for which exact solvers require hours to find and certify optimality.

The general idea of computing messages based on triangles is not novel in itself. 
It has been used for specific applications such as protein folding \citep{Jumper2021} and neural algorithmic learning \citep{ibarz2022a}. 
Furthermore, specific variants are formalized by simplicial \citep{bodnar21message} and hierarchical GNNs \citep{ritzert2019weisfeiler}.
The triangle messages we propose here are different: Firstly, they are defined with respect to edge features only. Secondly, each edge receives messages only from those pairs of edges with which it forms a triangle.
In particular, features or messages corresponding to nodes or structures like triangles and subgraphs are not considered. 
The experimental results demonstrate that, on small- and medium-sized instances, this adapted architecture and a simple supervised learning procedure are sufficient to outperform state-of-the-art heuristic solvers, and GNNs with standard architectures and more sophisticated training procedures.
\section{Related Work}
\label{sec:related-work}

\paragraph{Exact Solvers:}
Branch-and-cut algorithms are the most commonly used exact solvers for the multicut problem.
Although they have exponential worst-case time complexity, they can, depending on the cost structure, solve small- and medium-sized instances in feasible time \citep{kappes2011globallyoptimal, andres2012globally}.
Branch-and-cut algorithms operate by solving linear programming (LP) relaxations, and adding cutting planes or branching on fractional variables to obtain an optimal integer solution.
Although they are implemented by generic solvers, such as Gurobi \citep{gurobi} and CPLEX \citep{cplex}, they can be accelerated by adding problem-specific cutting planes, particularly those corresponding to facet-defining inequalities \citep{groetschel1989cutting, oosten2001clique, sorensen2020separation,letchford2024separation}.
Recently, \citet{irmai2025cutting} implemented such a specialized branch-and-cut algorithm for the clique partitioning problem \citep{groetschel1990facets} using Gurobi as underlying LP solver.
Their algorithm considers cutting planes for classes of inequalities that are not considered by any of the other solvers.
It is also the most recent publicly available exact solver for the multicut (or clique partitioning) problem, we are aware of.

\paragraph{Traditional Heuristic Solvers:}
Most heuristic multicut solvers are based on iteratively improving an initial solution through cost-reducing transformations.
The Kernighan and Lin algorithm with joins \citep{keuper2015efficient} searches locally for sequences of cost-reducing moves and joins. 
The greedy additive edge contraction (GAEC) algorithm \citep{keuper2015efficient} contracts edges with the highest positive cost until all edge costs become non-positive. 
The greedy fixation algorithm \citep{levinkov2017comparative} additionally fixes edges with large negative cost to be cut.
The Mutex Watershed algorithm \citep{wolf2018mutex} works similarly, but updates the costs after a contraction by taking the maximum rather than the sum.
The Cut Glue \& Cut algorithm \citep{beier2014cut} partitions and joins clusters of a given solution based on max-cut objectives.
It is generalized by the fusion moves algorithm \citep{beier2015fusion}, which generates a proposal solution and fuses it with the current solution by solving the multicut problem instances obtained by contracting all edges joined in both solutions.

Although the presented approaches can handle large instances infeasible for exact solvers, they perform transformations only sequentially and require storing the edge costs in memory, which are potentially quadratic in the number of nodes.
For very large instances, where sequential processing is no longer feasible, \citet{abbas2022rapid} introduce the rapid multicut algorithm. This primal-dual, GPU-based algorithm simultaneously contracts edges and optionally uses message passing on a Lagrangian decomposition \citep{swoboda2017message} to update edge costs.
Furthermore, to reduce the space requirements, \citet{abbas2023clusterfug} adapt the GAEC algorithm for instances whose costs are given as inner products of node features, such that only these node features must be stored.

\paragraph{GNN-based Heuristic Solvers:}
\citet{jung2023learning} train a GNN to predict the probability of edges being cut in an optimal solution to the multicut problem and round these to a feasible solution.
The GNN is trained using supervised learning with a binary cross-entropy loss and an additional term that encourages the feasibility of the predicted solutions.
Their approach is designed to solve large instances quickly, being significantly faster than traditional heuristic solvers like GAEC, at the cost of solution quality.
\citet{li2025dgrl} model the multicut problem as a Markov decision process, in which actions determine which edges to contract and states are learned by a GNN.
After using $Q$-learning to obtain an edge selection policy, they apply this policy to iteratively contract edges.
Their approach achieves high solution quality on the small- and medium-sized instances considered in their experiments, outperforming solvers like KL and GAEC.
However, they train separate models for each test dataset and use ensemble inference, which increases both training and inference time.

The approach, we present in this article combines elements of both works and introduces new ones.
Similar to \citet{jung2023learning}, we train our GNN supervised to predict which edges to cut and join in an optimal solution.
Similar to \citet{li2025dgrl}, we use an iterative inference procedure to contract edges in an autoregressive manner, and focus on obtaining high-quality solutions for small- and medium-sized instances.
Different from both, we only operate on edge features and update them using triangle-based message passing layers, which we design specifically for the multicut problem.

\section{Preliminaries}
\label{sec:preliminaries}

\subsection{Multicut Problem}

\paragraph{Multicuts:} 
Let $G = (V, E)$ be a graph.
A \emph{clustering} of $G$ is a partition $\Pi$ of $V$ such that for any $U \in \Pi$, any distinct $i, j \in U$ are connected in $G[U]$.
A set of edges $M \subseteq E$ is called a \emph{multicut} of $G$ if and only if there is a clustering $\Pi$ of $G$ such that $M$ consists precisely of those edges that straddle distinct clusters of $\Pi$.
Chopra and Rao \citep{chopra1993partition} show that there exists a one-to-one correspondence between the multicuts and the clusterings of a graph. 
In particular, $\phi_G(\Pi) = \{ij \in E \mid \forall U \in \Pi: \{i,j\} \not\subseteq U\}$ is a bijection from the clusterings to the multicuts of $G$.

\begin{figure}[t]
    \centering
    \begin{tikzpicture}[scale=1.6,base]
    \def \r {0.2cm}

    \node[style=vertex] (0) at (0.0, 1.1) {};
    \node[style=vertex] (2) at (1.7, 1.0) {};
    \node[style=vertex] (3) at (0.2, -0.1) {};
    \node[style=vertex] (4) at (0.9, 0.1) {};
    \node[style=vertex] (5) at (2.2, -0.4) {};
    \node[style=vertex] (6) at (2.9, 0.6) {};
    \node[style=vertex] (7) at (3.0, 0.0) {};

    \draw[dotted] (0) -- (2) node [midway, above] {\color{black}$-1$};
    \draw [dotted](0) -- (3) node [midway, left]  {\color{black}{$-4$}};;
    \draw (2) -- (4) node [midway, below, xshift=-0.85em, yshift=0.85em] {{$3$}};
    \draw[dotted] (2) -- (5) node [midway, right, yshift=0.8em, xshift=-0.3em] {\color{black}{$-5$}};
    \draw(3) -- (4) node [midway, below] {\color{black}{$1$}};
    \draw[dotted] (4) -- (5) node [midway, above, yshift=0.1em] {\color{black}{$2$}};
    \draw (5) -- (7) node [midway, below, xshift=0.3em] {{$-1$}};
    \draw (6) -- (7) node [midway, right] {{$2$}};
    \draw (5) -- (6) node [midway, right, xshift=0.1em, yshift=-0.1em] {{$4$}};
    \draw[dotted] (6) -- (2) node [midway, above, yshift=0.1em]{{$2$}};

    \path[fill=mygreen,draw=mygreen,opacity=0.3,line width=6ex,line cap=round,line join=round]
        (3)--(4)--(2)--cycle;

    \path[fill=mygreen,draw=mygreen,opacity=0.3,line width=6ex,line cap=round,line join=round]
        (5)--(7)--(6)--(5);

    \path[fill=mygreen,draw=mygreen,opacity=0.3,line width=6ex,line cap=round,line join=round]
        (0)--(0);
\end{tikzpicture}
    \caption{
        Depicted above is an example of an instance of the multicut problem.
        The dotted edges form an optimal multicut with cost $-6$.  
        The clustering induced by it is indicated by the shaded areas.}
    \label{fig:example-multicut}
\end{figure}
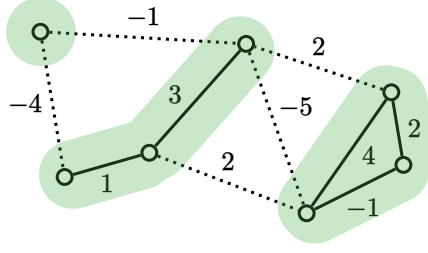

\paragraph{Problem Definition:} 
Given a graph $G = (V, E)$ with edge costs $c \in \mathbb{R}^E$, the multicut problem seeks a multicut that minimizes the cost of the cut edges (see \Cref{fig:example-multicut}).
It is formulated using binary variables $x \in \{0,1\}^E$, which indicate for every edge $e \in E$ if it is cut, $x_e = 1$, or joined, $x_e = 0$:
\begin{definition}
For any graph $G = (V, E)$,
any $c \in \mathbb{R}^E$
and $\MC_{G} \coloneqq \{ x \in \{0,1\}^E \mid x^{-1}(1) \text{ is a multicut of } G \}$,
we call $\min \{ \langle c, x \rangle \mid x \in \MC_{G} \}$
the instance of the multicut problem w.r.t.~$G$ and $c$.
\end{definition}
The feasible solutions of the multicut problem, i.e. the characteristic vectors of multicuts, can be characterized by a system of linear inequalities w.r.t. to the chordless cycles of the underlying graph:
\begin{lemma}[{\citealt{chopra1993partition}}]
\label{lemma:multicut-inequalities}
For any graph $G = (V, E)$ and any $x \in \{0,1\}^E$, $x^{-1}(1)$ is a multicut of $G$ if and only if
\begin{align}
\forall C \in \text{chordless-cycles}(G)
\ \forall e \in C \colon \quad 
x_e \leq\sum_{ e' \in C \setminus \{e\}} x_{e'}\,.
\label{eq:constraint-cycle}
\end{align}
\end{lemma} 

\paragraph{Graph Completion:}
The following lemma shows that, an instance of the multicut problem can be transformed into an equivalent instance on a complete graph by adding missing edges with cost $0$.
The chordless cycles of the resulting complete graph are precisely its triangles. 
Thus, the potentially exponential number of chordless cycle inequalities \eqref{eq:constraint-cycle} reduces to $3\tbinom{\lvert V \rvert}{3}$-many triangle inequalities:
\begin{lemma}[{\citealt{chopra1993partition}}]
\label{lemma:complete-graph}
Let $G = (V, E)$ be a graph and let $c \in \mathbb{R}^E$.
Let further $G' = (V, E')$ be the complete graph with $E' = \tbinom{V}{2}$ and let $c' \in \mathbb{R}^{E'}$ such that $c'_e = c_e$ if $e \in E$ and $c'_e = 0$ otherwise.
For any feasible solution $x'$ to the instance of the multicut problem w.r.t. $G'$ and $c'$, the restriction $x = x'|_E$ is a feasible solution to the instance w.r.t. $G$ and $c$.
Moreover, $x'$ is optimal if and only if $x$ is optimal.
\end{lemma}

\paragraph{Edge Contraction:} 
Many heuristic solvers for the multicut problem are based on edge contractions \citep{beier2015fusion, keuper2015efficient, abbas2022rapid, abbas2023clusterfug, li2025dgrl}.
In particular, these solvers iteratively contract edges while updating the corresponding costs until a termination criterion is reached.
The contraction of an edge $ij$ thereby corresponds to fixing nodes $i$ and $j$ to be in the same cluster, i.e. joining the edge $ij$ in the solution.
Once the termination criterion is reached, the remaining graph corresponds to a clustering of the original graph, and the remaining edges correspond to a multicut.
We now formalize this contraction operation for complete graphs:
\begin{definition}
    \label{def:contraction}
    For any complete graph $G = (V, E)$, any $c \in \mathbb{R}^{E}$ and any edge $ij \in E$, the graph and costs obtained by contracting $ij$ (and keeping $i$) are given by $G' = (V',E')$ with $V' = V \setminus \{j\}$, $E' = E \setminus \{e \in E \colon j \in e\}$ and $c' \in \mathbb{R}^{E'}$ such that $c'_{kl} = c_{ki} + c_{kj}$ if $l = i$, and $c'_{kl} = c_{kl}$ otherwise. 
\end{definition}

\subsection{Message Passing Neural Networks}
Graph neural networks \citep{scarselli2009graph} iteratively update node features based on the features of their neighbors and the features of the edges connecting them.
Message passing neural networks (MPNNs) \citep{gilmer2017neural} are a class of graph neural networks that implement these node updates in terms of a message passing scheme.
This scheme is implemented in message passing layers, which are the main building blocks of MPNNs. 
In the following, we introduce the original formulation of message passing layers, which has since been extended \citep{ritzert2019weisfeiler, bodnar21message}.

Let $G=(V,E)$ be a graph with initial node features $\{h_i \in \mathbb{R}^{d_\text{in}} \, \mid \, i \in V\}$ and initial edge features $\{h_{ij} \in \mathbb{R}^{d_\text{in}}  \, | \, ij \in E\}$ for some ${d_\text{in}} \in \mathbb{N}$. Let  further ${d_\text{out}} \in \mathbb{N}$ be the dimension of the updated features.
In the original formulation of message passing layers, each node $i \in V$ receives messages $m_{ij} \in \mathbb{R}^{d_\text{in}}$ from its neighbors $j \in N_G(i)$ based on a message function $M\colon \mathbb{R}^{3{d_\text{in}}} \to \mathbb{R}^{d_\text{in}}$. 
The received messages are then aggregated to a single message $m_i \in \mathbb{R}^{d_\text{in}}$, which is used to compute updated node features $h'_i \in \mathbb{R}^{{d_\text{out}}}$ based on an update function $U\colon \mathbb{R}^{2{d_\text{in}}}\to \mathbb{R}^{{d_\text{out}} }$:

\begin{align}
    m_{ij} = M(h_i, h_j, h_{ij})\,, \quad m_i = \sum_{j \in N_G(i)} m_{ij}\,, \quad h'_i = U(h_i, m_i)\,.    
\end{align}

\section{Triangle Message Passing GNNs}
\label{sec:gnn}
To solve a given instance of the multicut problem w.r.t. a graph $G=(V,E)$ and edge costs $c \in \mathbb{R}^E$, we train our model to predict logits for the edges to be contracted in an optimal solution.
We then use the model to compute a feasible solution by iteratively contracting the edges with the largest logit in an autoregressive manner.
In the following, we describe this process in detail, including preprocessing and training procedures.

\subsection{Preprocessing}

\paragraph{Graph Completion:} 
Feasible solution to the multicut problem are characterized by the chordless cycle inequalities \eqref{eq:constraint-cycle}.
These enforce that for every chordless cycle in the underlying graph, if an edge is cut, at least one other edge is also cut.
The number of chordless cycles can be exponential. Furthermore, for large chordless cycles, this information requires many message passing layers to propagate around the cycle.
To mitigate these problems, we complete the graph by adding missing edges with cost $0$ as described in \Cref{lemma:complete-graph}.
By this lemma, the instance w.r.t. to the resulting complete graph $G'=(V, E')$ and costs $c' \in \mathbb{R}^{E'}$ is equivalent to the original one.

\paragraph{Cost Normalization:}
It is easy to see that the multicut problem is invariant to scaling the costs by a positive constant.
To reflect this invariance, we consider normalized costs $\tilde{c}' \in \mathbb{R}^{E'}$ obtained by dividing the sum of absolute edge costs and multiplying by the number of edges:
\begin{align}
    \forall e \in E'\colon \quad \tilde{c}'_e = \frac{c'_e}{\sum_{e' \in E'} \lvert c'_{e'} \rvert }\lvert E' \rvert\,.
\end{align}
This normalization also ensures that the expected absolute cost of each edge is $1$, stabilizing the training of our model and improving its generalization across instances of varying size.

\subsection{Triangle Message Passing Layers}
%
From the preprocessing, we obtain a complete graph $G'=(V,E')$ with normalized costs $\tilde{c}' \in \mathbb{R}^{E'}$.
Thus, for any distinct nodes $i,j \in V$, there exists an edge $ij \in E'$.
Furthermore, any other distinct node $k \in V \setminus \{i,j\}$ forms a triangle with $i$ and $j$.

Since the objective function and constraints of the multicut problem are defined on edges, we assign features $h_{ij} \in \mathbb{R}^{d_\text{in}}$ only to edges $ij \in E'$, avoiding ambiguities.
For any distinct nodes $i,j,k \in V$, we define a message $m_{ijk} \in \mathbb{R}^{d_\text{in}}$ from edges $ik$ and $jk$ to $ij$. 
For a fixed $i$ and $j$, messages $m_{ijk}$ are then aggregated to a single message $m_{ij} \in \mathbb{R}^{d_\text{in}}$, which is used to compute the updated features $h_{ij}' \in \mathbb{R}^{d_\text{out}}$:
%
%
%
\begin{align}
    m_{ijk} = M\bigl(h_{ij}, h_{ik} + h_{jk}, \lvert h_{ik} - h_{jk} \rvert\bigr), \quad 
    m_{ij} = \frac{1}{\lvert V \rvert - 2} \sum_{k \in V \setminus \{i,j\}} m_{ijk}, \quad 
    h_{ij}' = U\bigl(h_{ij}, m_{ij}\bigr)\,. 
\end{align}
Functions $M\colon \mathbb{R}^{3d_\text{in}} \rightarrow \mathbb{R}^{d_\text{in}}$ and  $U\colon \mathbb{R}^{2d_\text{in}} \rightarrow \mathbb{R}^{d_\text{out}}$ are thereby multi-layer perceptrons with GELU \citep{hendrycks2023gelu} activation functions.

By construction, the messages $m_{ijk}$ are invariant under the permutation of $i$ and $j$. 
Furthermore, there exists a bijection between the message $m_{ijk}$ and the triangle inequalities that characterize feasible solutions to the multicut problem for complete graphs.
Thus, these messages directly capture the violation of triangle inequalities.
%

\subsection{Model Architecture}

Our model consists of $20$ consecutive triangle message passing layers.
We initialize the edge features $h_{ij} \in \mathbb{R}$ for each edge $ij \in E'$ by the normalized cost $\tilde{c}_{ij}'$. 
All intermediate layers have a hidden feature dimension of $64$ and incorporate layer normalization and residual connections. 
The output layer projects the features to a dimension of $1$ and omits the GELU activation function of $U$. 
We interpret the output as logits $z \in \mathbb{R}^{E'}$ for the edges to be contracted in an optimal solution.

With the specified number of triangle message passing layers and hidden feature dimension, our model has $403\,719$ learnable parameters.
Since the number of triangles in a complete graph is $3\tbinom{\lvert V \rvert}{3}$, a pass of the model has a time complexity and a space complexity of $\mathcal{O}(\lvert V \rvert^3)$.

\subsection{Inference}
We apply our model in an autoregressive manner to heuristically solve the multicut problem.
After preprocessing, we use our model to obtain edge logits $z \in \mathbb{R}^{E'}$.
We then select the edge $e \in \argmax_{e' \in E'} z_{e'}$ with the highest logit and contract it, updating the graph and the costs according to \Cref{def:contraction}.
We repeat this process until all edge logits are non-positive.
Finally, we return the obtained feasible solution.
This autoregressive approach increases the worst-case time complexity to $\mathcal{O}(\lvert V \rvert^4)$, but significantly improves the solution quality compared to a single pass as shown in Section \ref{sec:experiments-ablation}.
Like other heuristic solvers, this approach provides no approximation guarantees.

\subsection{Training}
\label{sec:gnn-training}
We train our model supervised, on synthetic data.
For each combination of graph size $n \in \{10, 15, 20, 25, 30\}$ and costs sampled uniformly from integers in the range $r \in \{[-1,1], [-5,5], [-100,100]\}$, we generate $10\,000$ instances of the multicut problem.
Thus, our training set consists of $150\,000$ instances in total.
We then compute optimal solutions for these instances using an exact branch-and-cut algorithm \citep{irmai2025cutting}.
%
%
When sampling an instance from the training set, we further augment it by randomly contracting edges not cut in the optimal solution. 
This yields multiple training signals from the computation of a single optimal solution and further improves the generalization of our model.

As loss, we consider the mean binary cross-entropy between the logits $z \in \mathbb{R}^{E'}$ after applying the sigmoid function $\sigma\colon \mathbb{R} \rightarrow [0,1]$, and the optimal solution $x^* \in \{0,1\}^{E'}$:
\begin{align}
    \mathcal{L}(z, x^*) = -\frac{1}{\lvert E' \rvert} \sum_{e \in E'} 
    \Bigl(x^*_e \log(\sigma(z_e)) + (1 - x^*_e) \log(1 - \sigma(z_e))\Bigr)\,.
\end{align}
We train our model for $500$ epochs using the Adam optimizer \citep{kingma2015adam}, with a cosine annealing learning rate ranging from $10^{-4}$ to $10^{-6}$ and a batch size of $1$.
\section{Experiments}
\label{sec:experiments}
In this section, we evaluate our model against heuristic and exact solvers using synthetic and real-world data, and provide an ablation study. 

\subsection{Datasets and Experimental Setup}
\label{sec:experiments-setup}
We primarily evaluate our model on the CP-Lib benchmark of \citet{sorensen2024cplib}.
This benchmark comprises a diverse set of synthetic and real-world instances.
The instances range in size from $30$ to $2\,500$ nodes and are categorized into seven datasets based on their origin and characteristics.
Not all instances of the benchmark have known optimal solutions.
In these cases, the best known solutions from literature are reported.
Due to the cubic time and space complexity of a model pass, we only consider instances with up to $200$ nodes in our experiments.
This results in a total of $152$ instances from the CP-Lib benchmark for evaluation.
To test the generalization capability of our model with respect to graph size, we additionally consider a dataset of random instances generated analogously to the training data.
In particular, this dataset consists of $100$ instances for each graph size $n \in \{10,20,\dots,200\}$ with costs sampled uniformly from integers in range $[-5,5]$.

All experiments are performed on a machine with an Intel Core i9-12900KF CPU @ 5.20\,GHz and an NVIDIA GeForce RTX 4080 Super GPU.
Training with this setup and the procedure described in Section \ref{sec:gnn-training} has taken $12$ hours.
If an optimal solution is available, we evaluate the quality of a given solution by its optimality gap.
The optimality gap is defined as $\frac{c(x) - c(x^*)}{\lvert c(x^*) \rvert}$, where $c(x)$ is the objective value of the obtained solution, and $c(x^*)$ is the optimal objective value.
For comparability, we solve each instance separately and do not parallelize over instances.

\subsection{Heuristic Solvers} 
We compare our model against the greedy additive edge contraction algorithm (GAEC) \citep{keuper2015efficient}, the Kernighan and Lin algorithm with joins (KL) \citep{keuper2015efficient}, and the fusion moves algorithm (FM) \citep{beier2015fusion} as implemented by the \href{https://github.com/DerThorsten/nifty}{nifty library (MIT license)}. 
Furthermore, we compare against the deep graph reinforcement learning (DGRL) approach of \citet{li2025dgrl}, who make their code \href{https://github.com/lizhenchen/DGRL_solve_multicut}{publicly available}.
We do not consider the solvers of \citet{abbas2022rapid,abbas2023clusterfug} and \citet{jung2023learning}, since they focus on solving large instances quickly, obtaining similar or worse objective values than GAEC for the instances considered in these articles.
In contrast, our solver is designed to solve small- and medium-sized instances close to optimality.
For FM, we use watershed-based proposals \citep{wolf2018mutex} and the Kernighan and Lin algorithm with joins to solve the contracted instances.
For DGRL, we use an ensemble of $10$ models trained on the same data as our model.

\begin{table*}[t]
\centering
\setlength{\tabcolsep}{2pt}
\caption{Mean optimality gaps and runtimes for solving datasets of the CP-Lib benchmark with up to $200$ nodes with heuristic solvers.} 
\label{tab:approximate-solvers}
\begin{tabular}{l | ccccc | ccccc}
\toprule
\multicolumn{1}{c|}{\multirow{2}{*}{\textbf{Dataset}}} & \multicolumn{5}{c|}{\textbf{Optimality Gap [$10^{-3}$]}} & \multicolumn{5}{c}{\textbf{Runtime [s]}} \\ 
& \multicolumn{1}{c}{Ours} & \multicolumn{1}{c}{KL} & \multicolumn{1}{c}{FM} & \multicolumn{1}{c}{GAEC} & \multicolumn{1}{c|}{DGRL} & \multicolumn{1}{c}{Ours} & \multicolumn{1}{c}{KL} & \multicolumn{1}{c}{FM} & \multicolumn{1}{c}{GAEC} & \multicolumn{1}{c}{DGRL} \\
\midrule
ABR         & 3.44           & \phantom{0}\bfseries 0.01 & 85.91 & 91.22 & {78.26} & 13.10 & 0.0028 & 0.0348 & \bfseries 0.0020 & {2308.85} \\
Artificial  & \bfseries 0.00 & \phantom{0}0.22           & \phantom{0}0.22  & \phantom{0}0.22  & {\phantom{0}0.21} & 25.78 & 0.0040 & 0.0998 & \bfseries 0.0020 & {6370.35} \\
ClusEdit    & \bfseries 1.43 & 33.43          & 61.27 & 85.99 & {66.02} & \phantom{0}1.42 & 0.0011 & 0.0182 & \bfseries 0.0006 & \phantom{00}{78.52} \\
Correlation \, & \bfseries 6.21 & 18.22          & 52.45 & 64.31 & {37.07} & \phantom{0}1.20 & 0.0011 & 0.0088 & \bfseries 0.0006 & \phantom{00}{28.20} \\
Equicut     & \bfseries 4.62 & 33.44          & 38.05 & 67.09 & {46.74} & \phantom{0}0.64 & 0.0004 & 0.0076 & \bfseries 0.0003 & \phantom{000}{6.92} \\
MCF         & \bfseries 1.19 & \phantom{0}5.16           & \phantom{0}5.35  & 14.30 & {24.10} & \phantom{0}1.94 & 0.0004 & 0.0076 & \bfseries 0.0003 & \phantom{00}{20.25} \\
Random      & 1.27           & \phantom{0}\bfseries 0.65 & \phantom{0}0.68  & \phantom{0}1.01  & \phantom{0}{0.95} & \phantom{0}1.86 & 0.0013 & 0.0090 & \bfseries 0.0007 & \phantom{000}{7.55} \\
\hline
\end{tabular}
\end{table*}

\Cref{tab:approximate-solvers} shows the results of the experiments on the CP-Lib benchmark.
The numbers indicate mean values for the datasets.
Only instances of the datasets for which an optimal solution is known are considered when computing the mean.
Plots showing quantiles and per-instance results are provided in \Cref{appendix:figures}.
\Cref{appendix:table} contains a table with results for all instances of the CP-Lib benchmark with up to $200$ nodes, including those for which no optimal solution is known.
In this table, we report the best known objective value from the literature, the objective value of the solution computed by our model and its runtime.

As evident from \Cref{tab:approximate-solvers}, our model significantly outperforms the considered heuristic solvers in terms of optimality gaps for all but the ``ABR'' and ``Random'' dataset. 
The reason for the worse performance of our model on these datasets might be related to the fact that their instances require more iterations to be solved, potentially leading to error accumulation. 
In terms of runtimes, our model is slower than GAEC, KL and FM on all considered datasets, but faster than DGRL.
This is expected due to the quartic time complexity of our model.
Still, its runtimes are in the order of seconds. 
The relative performance of DGRL compared to the other solvers is worse than what was originally reported by \citet{li2025dgrl}.
We believe that this is due to the hardness of the benchmark instances and insufficient generalization from out-of-distribution training data.

\begin{figure}[t]
    \centering
    \includegraphics[height=0.35\textwidth]{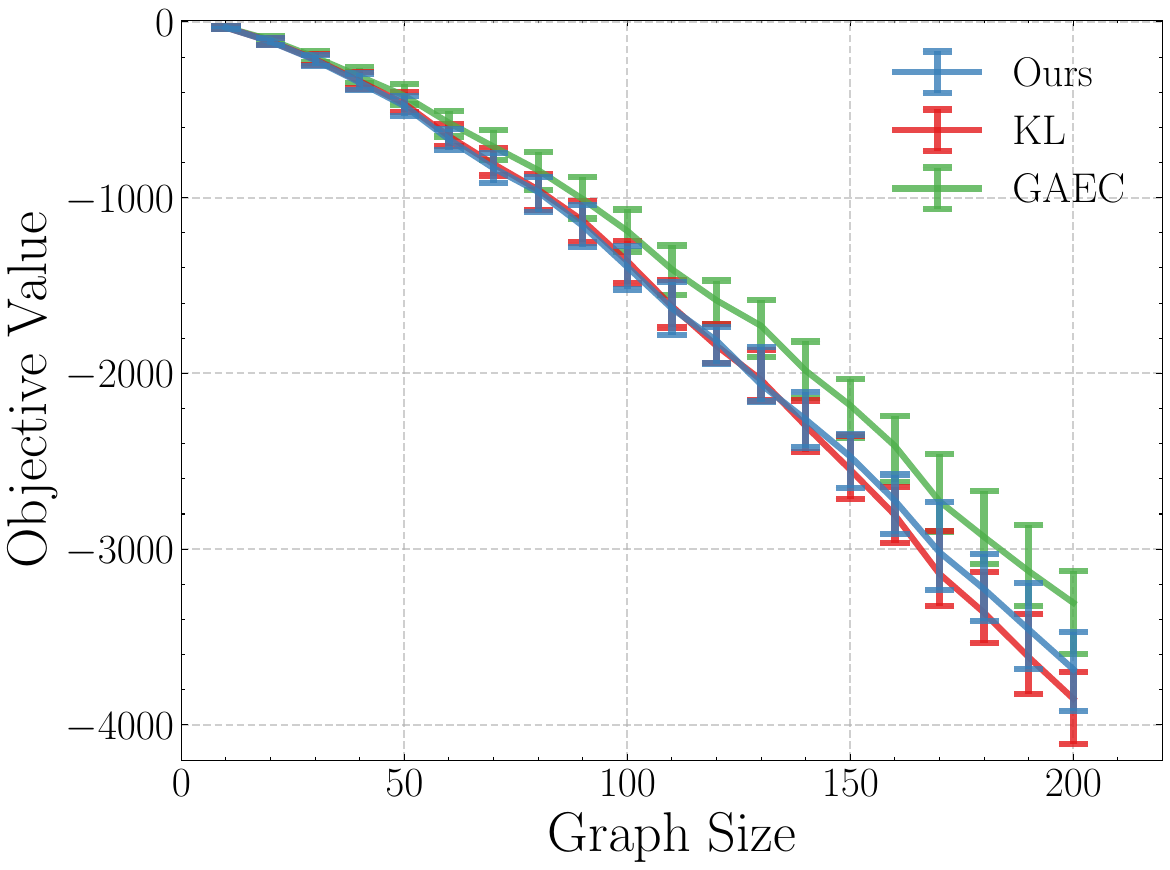}
    \quad
    \includegraphics[height=0.35\textwidth]{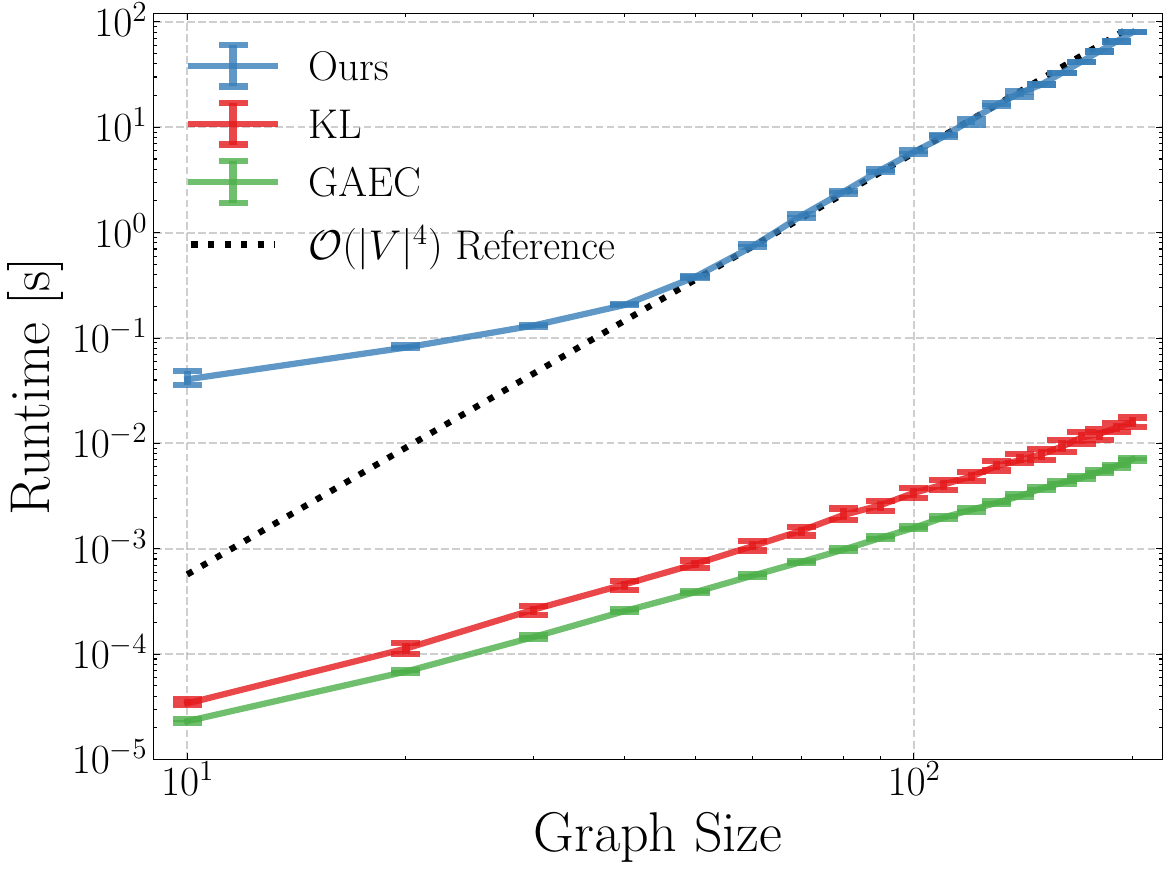}
    \caption{
        Depicted on the left are the median objective values of our model, KL and GAEC as a function of the graph size (number of nodes) for the dataset of random instances.
        Depicted on the right is the median runtime for these solvers on the same dataset.
        The error bars indicate the $0.25$- and $0.75$-quantile.
        The plot on the right uses a double logarithmic scale.
        The dotted line indicates the expected quartic growth of the runtime of our model.
    }
    \label{fig:large-random-instances}
\end{figure}
\Cref{fig:large-random-instances} shows median objective values and runtimes over the graph size for the dataset of random instances generated analogously to the training data.
Given the results from \Cref{tab:approximate-solvers}, we only show values for our model, KL and GAEC.

As the plot on the left shows, our model achieves better optimality gaps than KL for graph sizes up to around $140$.
For larger graphs, its performance degrades, but still surpasses GAEC.
This decrease is expected since our model is only trained on instances with up to $30$ nodes, for which structures relevant to larger instances cannot occur.
As seen in the plot on the right, our model is slower than KL and GAEC and exhibits the expected quartic growth in runtime.

\subsection{Exact Solvers}
In \Cref{tab:exact-solvers}, we compare the runtimes of our model to the runtimes of the exact branch-and-cut algorithm of \citet{irmai2025cutting} on selected instances of the CP-Lib benchmark, that our model solves to optimality.
\begin{table}[t]
\setlength{\tabcolsep}{3pt}
\centering
\caption{Runtimes of our model and the exact branch-and-cut solver of \citet{irmai2025cutting} on selected instances of the CP-Lib benchmark, that our model solves to optimality. We provide both, the time needed by the exact solver to find the optimal solution (B\&C) and the time needed to certify optimality ($\text{B\&C}^*$).}
\label{tab:exact-solvers}
\begin{tabular}{l | c c c c }
\toprule
\multicolumn{1}{c|}{\multirow{2}{*}{\textbf{Instance}}} & \multicolumn{3}{c}{\textbf{Runtime [}s\textbf{]}} \\ & Ours & B\&C & $\text{B\&C}^*$ \\
\midrule
cars & \phantom{0}0.18 & \phantom{000}0.008 & \phantom{000}0.016 \\
companies \, & 20.64 & \phantom{000}0.039 & \phantom{000}0.117 \\
corr60-3 & \phantom{0}0.81 & \phantom{000}0.008 & \phantom{0}252.898 \\
neg-c-70 & \phantom{0}0.44 & \phantom{00}30.148 & \phantom{00}30.148 \\
ce50-40 & \phantom{0}0.47 & 8064.828& 8064.828 \\
\hline
\end{tabular}
\end{table}

There is no correlation between the runtime of our model and that of the exact solver.
This is to be expected, since the runtime of our model depends on the size of the graph and the number of edges that need to be contracted to obtain an optimal solution.
In contrast, the runtime of the exact solver is dominated by the number of cutting planes that need to be added to the linear programming relaxation and the number of branching steps.
Thus, for instances like ``companies'', that have many nodes but an integer optimal solution in the linear programming relaxation, our model is slower than the exact solver.
On the other hand, for instances like ``ce50-40'', that have few nodes but are challenging for the exact solver, our model is significantly faster.
\Cref{appendix:figures} contains a figure visualizing the latent space of our model for the instances ``cars'',  ``ce50-40'' solved to optimality and the instance ``CPn35-3'' not solved to optimality.

\subsection{Ablation Study}
\label{sec:experiments-ablation}
\Cref{tab:ablation-study} shows the results of an ablation study evaluating the key components of our model. 
In the following, we briefly discuss each ablation. 

\begin{table*}[t]
\setlength{\tabcolsep}{2.5pt}
\centering
\caption{Mean optimality gaps and runtimes for solving datasets of the CP-Lib benchmark with up to $200$ nodes. We compare our model with ablated versions where the key components of graph completion (Completion), triangle-based message passing (TMP) and autoregressive inference (Inference) are removed.}
\label{tab:ablation-study}
\begin{tabular}{l | c c c c | c c c c}
\toprule
\multicolumn{1}{c|}{\multirow{2}{*}{\textbf{Dataset}}} & \multicolumn{4}{c|}{\textbf{Optimality Gap [$10^{-3}$]}} & \multicolumn{4}{c}{\textbf{Runtime [}s\textbf{]}} \\ & Ours & Completion & TMP & Inference & Ours & Completion & TMP & Inference\\
\midrule
ABR &3.44 & 14.76 & \phantom{0}6.78 &  \phantom{00}\textbf{0.33} & 13.10 & 13.77 & 17.07 &\textbf{0.393}\\
Artificial & \textbf{0.00} & \phantom{0}\textbf{0.00} & \phantom{0}0.38 & \phantom{00}0.10 & 25.78 & 25.78 & 28.05 &\textbf{0.691}\\
ClusEdit & \textbf{1.43} & \phantom{0}2.93 & 11.50 & \phantom{0}94.82 & \phantom{0}1.42 & \phantom{0}1.25& \phantom{0}1.36 &\textbf{0.076}\\
Correlation \, & \textbf{6.21} & \phantom{0}7.39 & 31.05 & \phantom{0}36.80 & \phantom{0}1.20 & \phantom{0}1.16 & \phantom{0}1.27 &\textbf{0.074}\\
Equicut & \textbf{4.62} & \phantom{0}6.67 & 24.73 & 158.97 & \phantom{0}0.64 & \phantom{0}0.56& \phantom{0}0.56 &\textbf{0.068}\\
MCF & \textbf{1.19} & \phantom{0}4.42 & \phantom{0}9.67 & \phantom{0}14.58 & \phantom{0}1.94 & \phantom{0}1.80 & \phantom{0}1.99 &\textbf{0.081}\\
Random & \textbf{1.27} & \phantom{0}1.47  & \phantom{0}2.49 & \phantom{00}5.78& \phantom{0}1.86 & \phantom{0}1.87 & \phantom{0}2.13 & \textbf{0.083}\\
\hline
\end{tabular}
\end{table*}

\paragraph{Graph Completion:}
We remove the graph completion step from our preprocessing and compute messages based on individual edges if they are no longer part of a triangle.
As shown in \Cref{tab:ablation-study}, this increases the optimality gaps on all datasets except ``Artificial''.
The effect is more pronounced for datasets containing sparse graphs such as ``ABR''.
It has no effect for the ``Artificial'' dataset, since all instances in this dataset are complete graphs.
The runtimes decrease slightly, since fewer edges need to be considered for edge contraction.
However, the time complexity of inference remain quartic in the number of nodes, since we do not adapt our architecture to sparse graphs.

\paragraph{Triangle Message Passing:}
We replace our triangle message passing layers by standard edge message passing layers, in which messages are computed based on neighboring edges instead of triangles.
Similar to the previous modification, this increases the optimality gaps on the datasets considered.
The effect is thereby more pronounced than in the case of omitting graph completion.
The runtimes show no significant change.

\paragraph{Autoregressive Inference:} 
We contract the edge with the largest logit, update the logits, and repeat this process until there are no positive logits remaining, without evaluating the GNN again.
This approach is equivalent to GAEC when logits are considered instead of edge costs for contraction.
As shown in \Cref{tab:ablation-study}, this modification significantly increases the optimality gaps for all but the ``ABR'' dataset. 
The improved optimality gaps for this dataset suggest again that error accumulation may be an issue for our model over the course of many iterations.
Only evaluating a single pass of the network reduces the time complexity of inference to $\mathcal{O}(\lvert V \rvert^3)$, leading to strongly reduced runtimes, as expected.
However, since both solution quality and runtime lag behind that of KL, this approach is not competitive.

\section{Conclusion}
\label{sec:conclusion}
In this article, we introduce a GNN-based heuristic solver for the multicut problem.
First, we preprocess the input graph by completing it and normalizing the edge costs.
Then, we apply triangle message passing layers that operate on edge features and compute messages based on triangles.
By construction, the messages computed in these layers correspond one-to-one to the triangle inequalities that define feasible solutions to the multicut problem on complete graphs.
We train our model using supervised learning with randomly generated instances to predict which edges should be contracted in an optimal solution.
We perform inference by iteratively contracting the edge with the largest logit in an autoregressive manner.
We empirically demonstrate the effectiveness of our approach by comparing it with heuristic and exact solvers on synthetic and real-world instances with up to 200 nodes. 
%
%
Like other heuristic solvers for the multicut problem, our approach does not provide any approximation guarantees.

One direction for future work is to adapt the training and inference procedures.
Currently, the size of the instances in our training set is limited by the need to compute optimal solutions.
Using reinforcement learning or self-training could allow scaling to larger training instances and potentially improve the performance of our model for larger instances.
Furthermore, the inference procedure could be adapted by using more advanced techniques, such as beam search, top-$k$ sampling, or top-$p$ sampling, instead of deterministically sampling the edge with the highest logit. 
%
%
Another direction for future work is to adapt our architecture to sparse graphs.
This could be accomplished by learning which edges to add during preprocessing, or by considering other graph structures, such as small chordless cycles, for message passing. 

\newpage

\bibliography{references}

\newpage

\appendix

\section{Additional Figures}
\label{appendix:figures}

\begin{figure}[h!]
    \centering
    \includegraphics[width=0.9\textwidth]{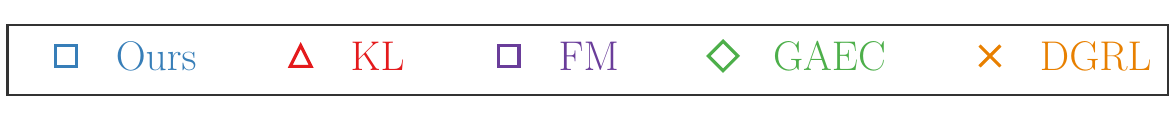}
    \vspace{12pt}
    \includegraphics[width=0.9\textwidth]{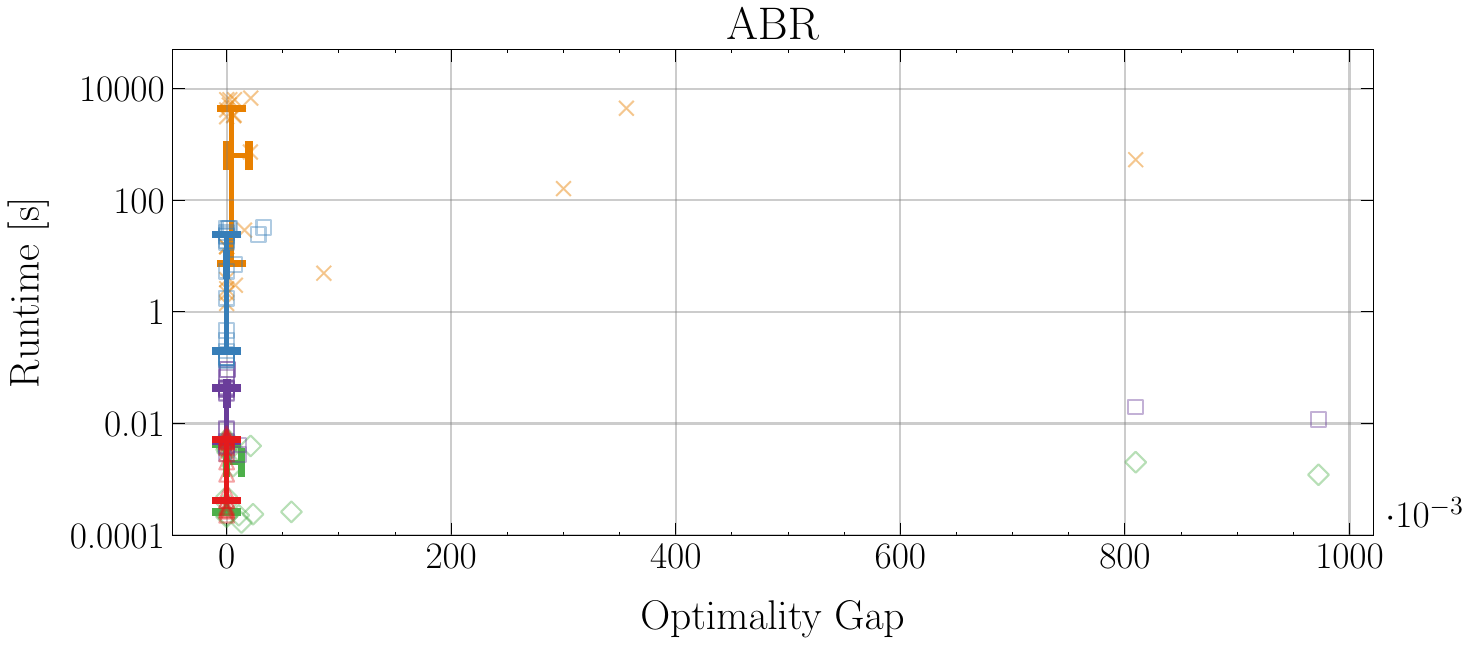}
    \vspace{12pt}
    \includegraphics[width=0.9\textwidth]{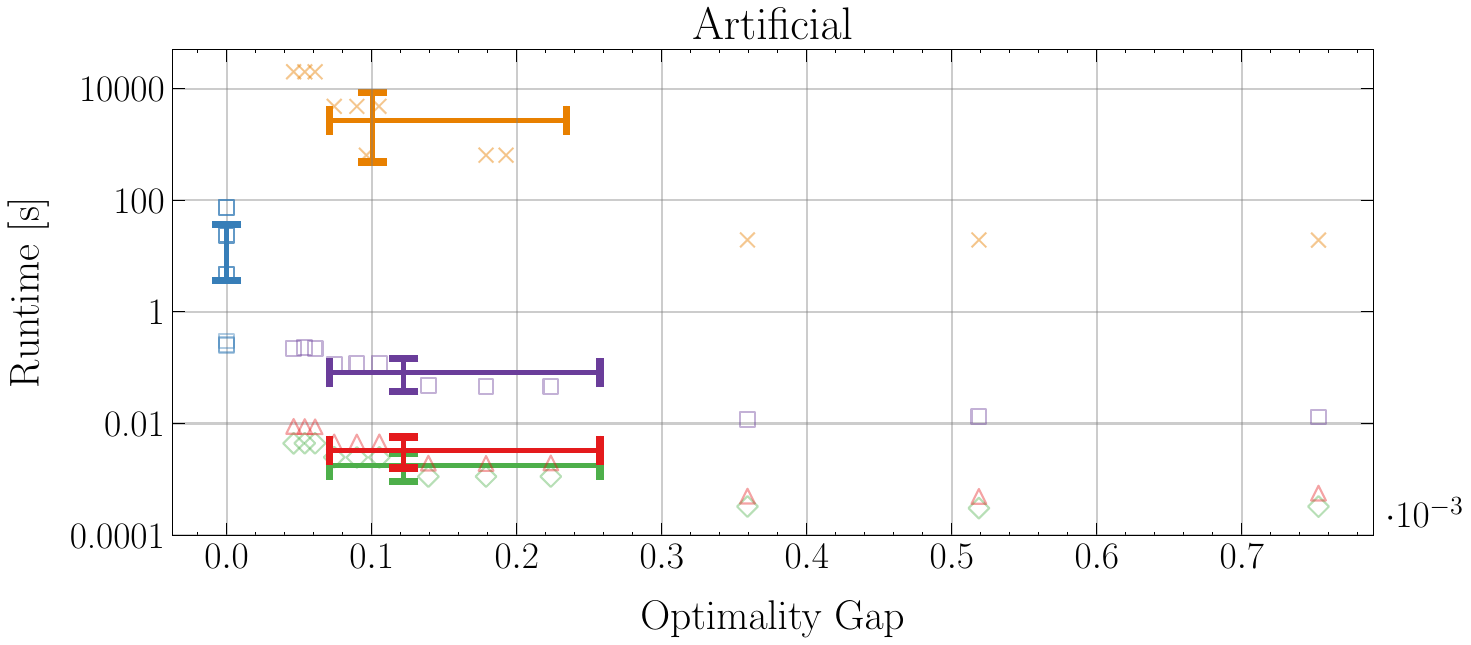}
    \vspace{12pt}
    \includegraphics[width=0.9\textwidth]{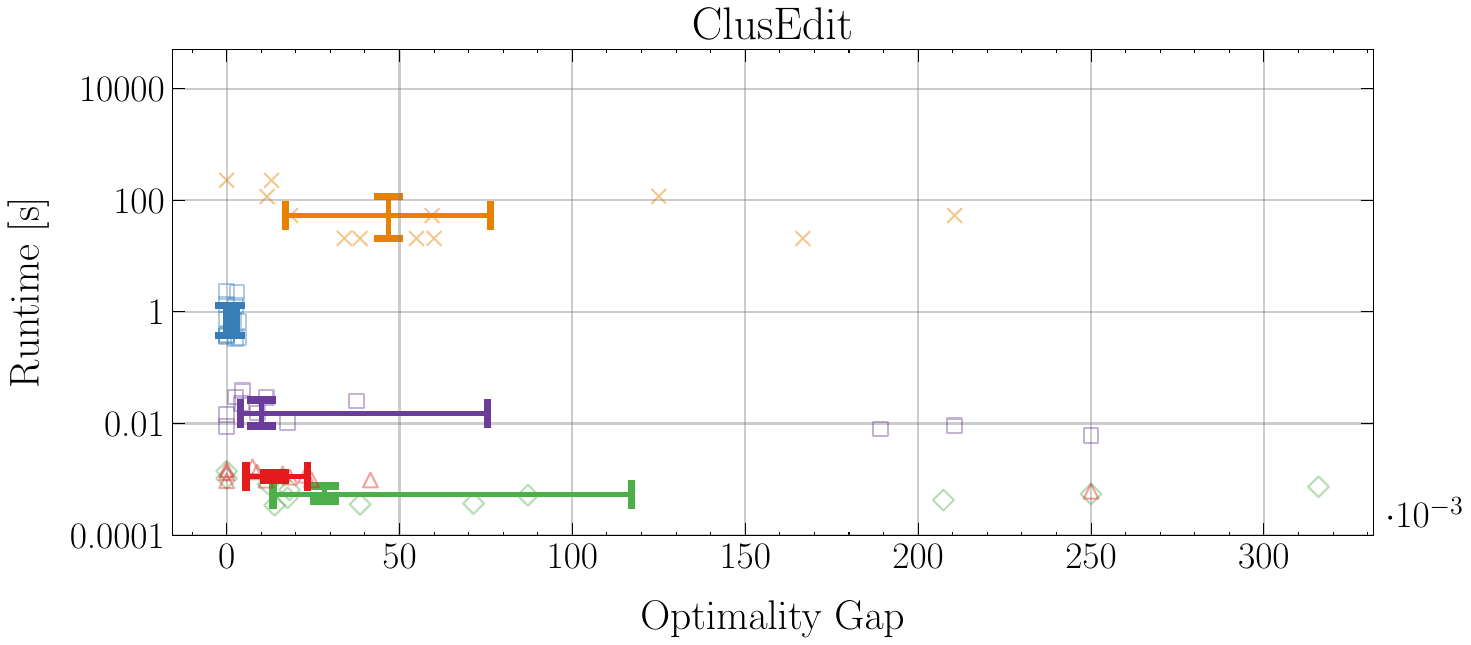}
    \vspace{12pt}
    \textit{(continued on next page)}
\end{figure}

\begin{figure}[h!]
    \centering
    \includegraphics[width=0.9\textwidth]{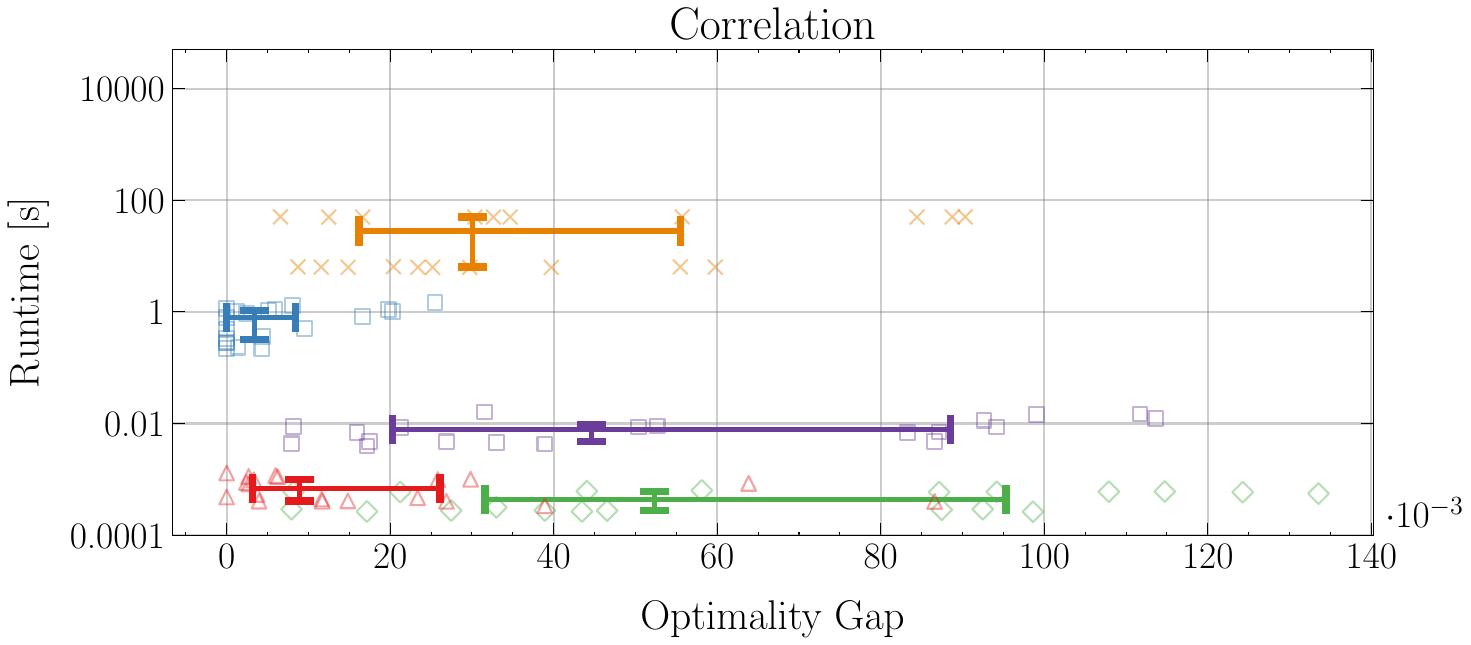}
    \vspace{12pt}
    \\
    \includegraphics[width=0.9\textwidth]{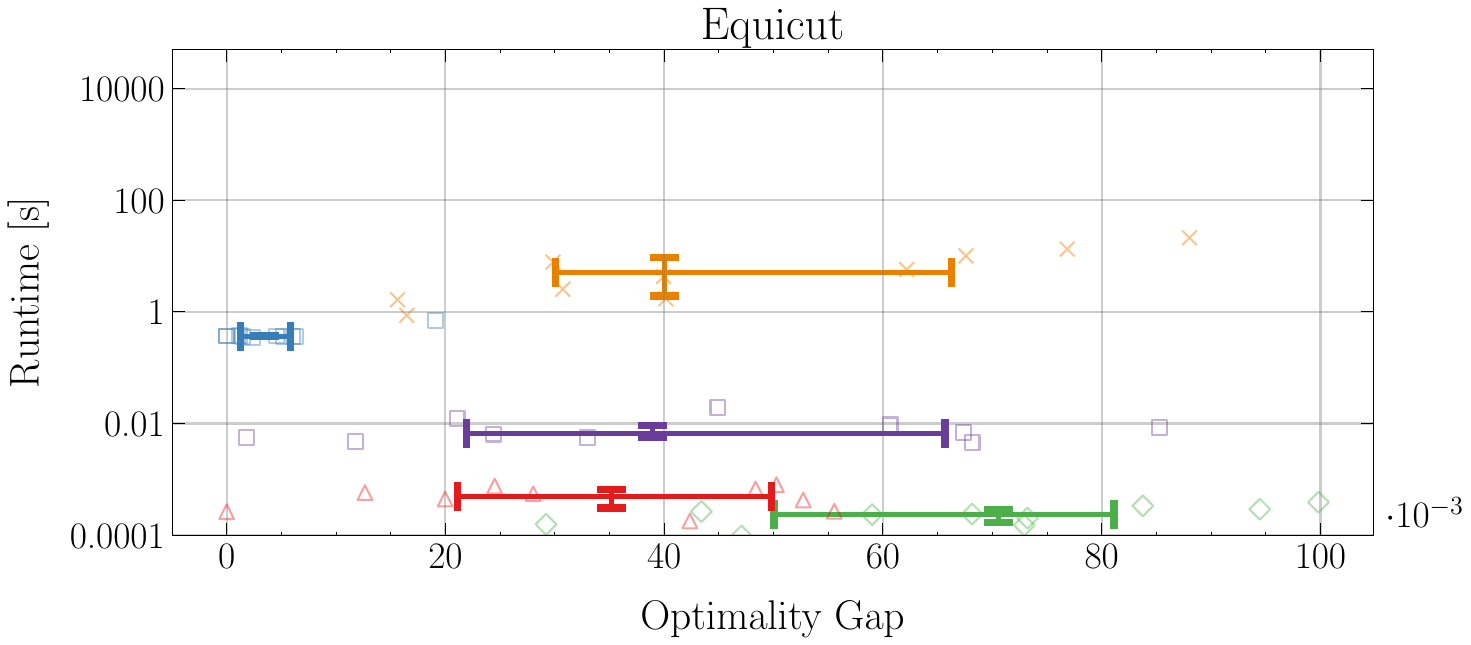}
    \vspace{12pt}
    \\
    \includegraphics[width=0.9\textwidth]{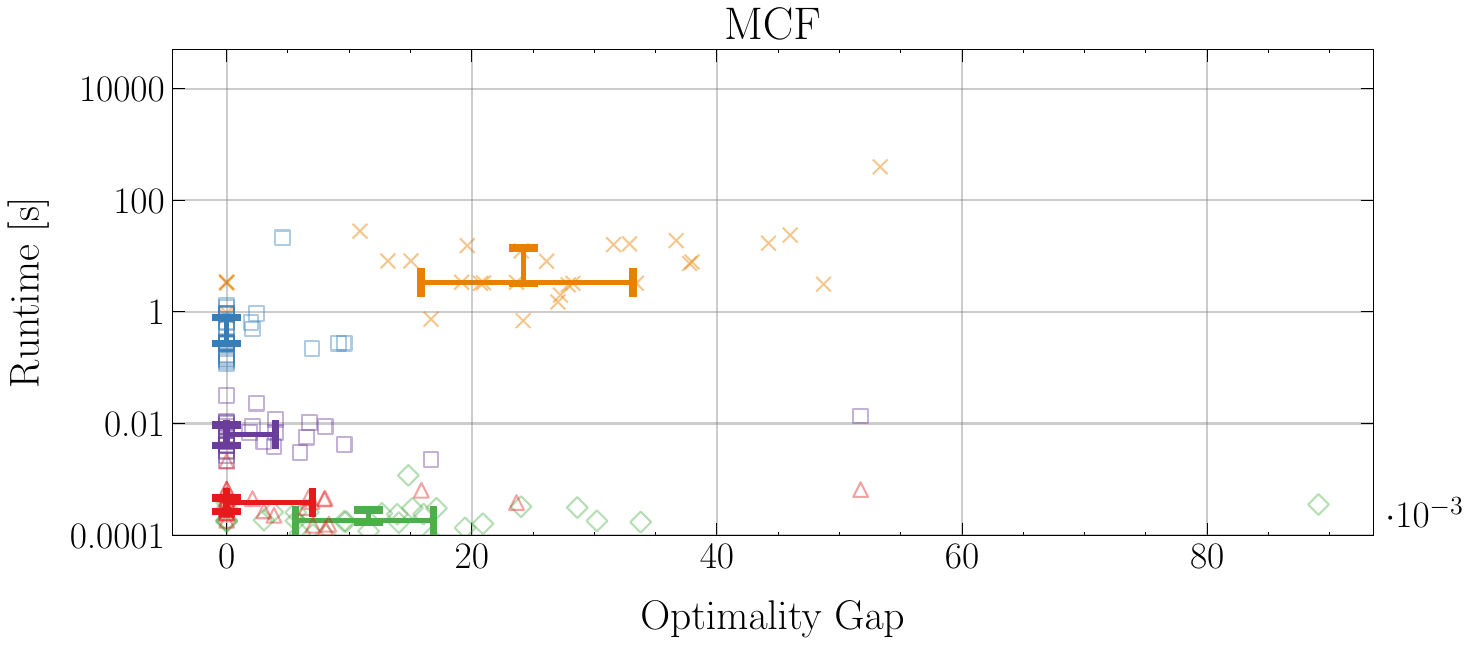}
    \vspace{12pt}
    \\
    \textit{(continued on next page)}
\end{figure}

\clearpage

\begin{figure}[!t]
    \centering
\includegraphics[width=0.9\textwidth]{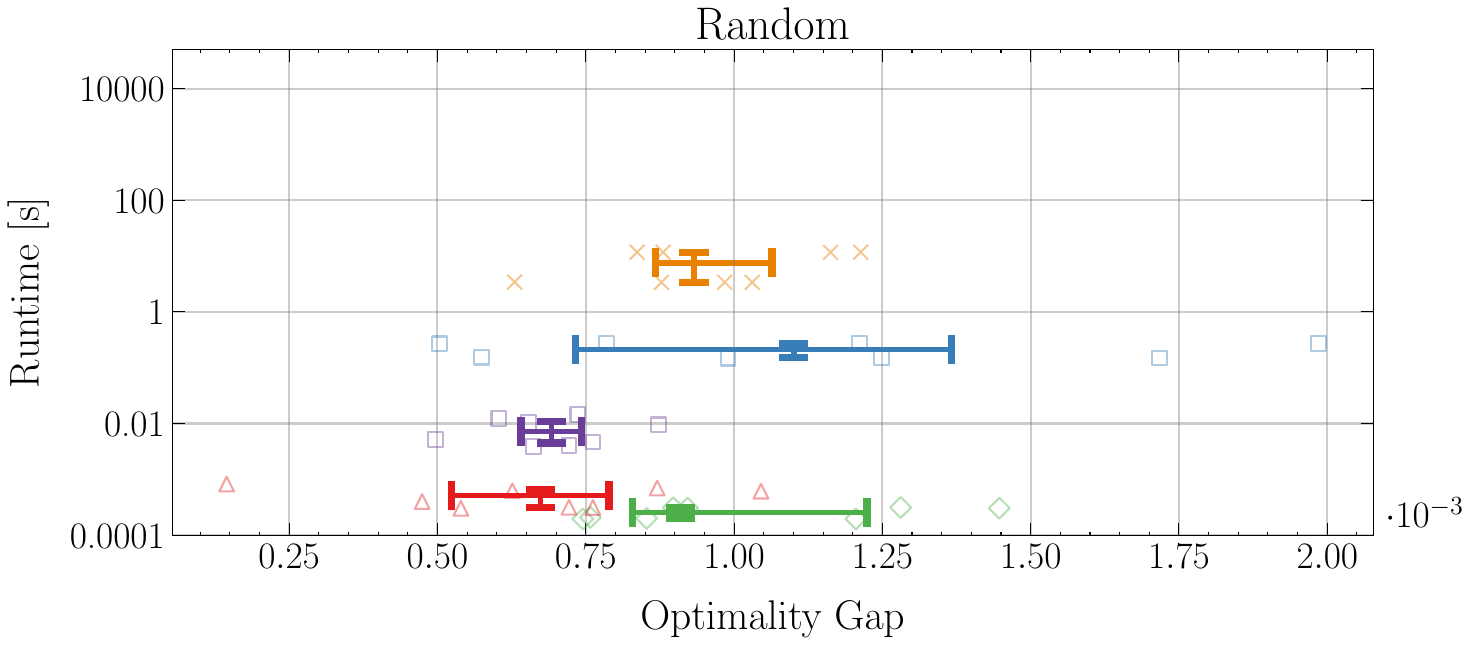}
    \caption{
        Optimality gaps and runtimes for solving instances of the CP-Lib benchmark \cite{sorensen2024cplib} with up to 200 nodes and known optimal solutions with our model, KL \cite{keuper2015efficient}, FM \cite{beier2015fusion}, GAEC \cite{keuper2015efficient} and DGRL \cite{li2025dgrl}.
        Each plot shows the instances of a different dataset. 
        The markers correspond to values for the individual instances, while the error bars show the $0.25$- and $0.75$-quantile.
        Their intersection corresponds to the median.
        For datasets ``ABR'', ``Artificial'' and ``MCF'', no error bars with respect to the optimality gap are shown for some solvers since the $0.25$- and $0.75$-quantile are both $0$. 
        Note the logarithmic scale for the runtime and different scales for the optimality gap.
    }
    \label{fig:cplib-quantiles}
\end{figure}

\vspace*{10cm}
    
\clearpage

\begin{figure}[h]
    \centering
    \includegraphics[height=0.45\textwidth]{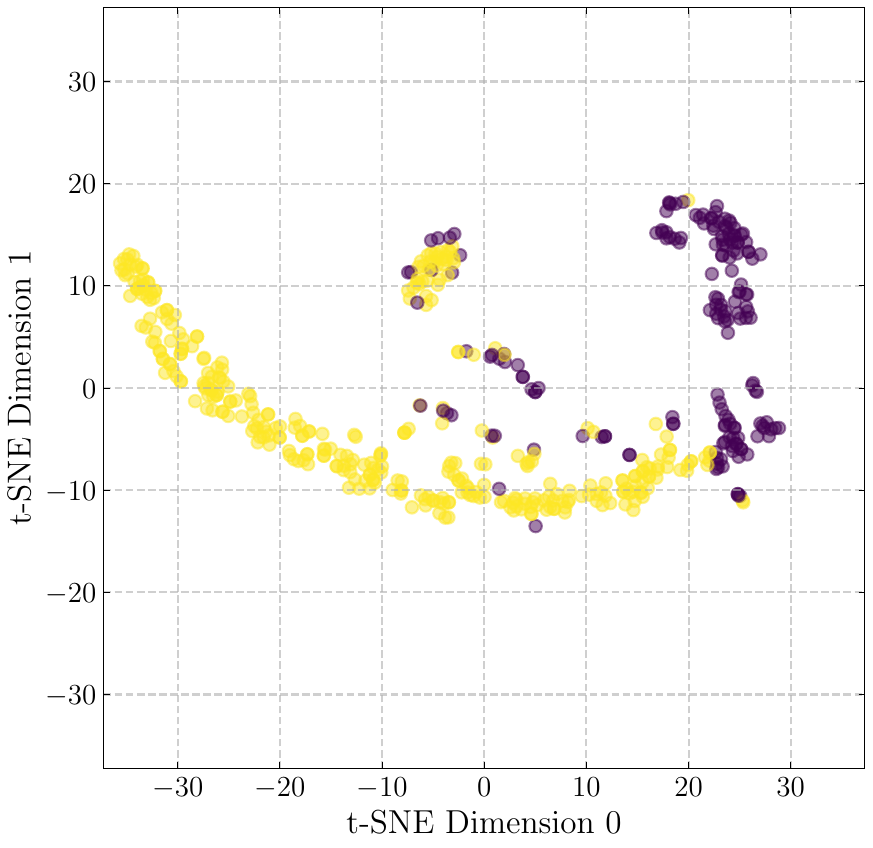}
    \includegraphics[height=0.45\textwidth]{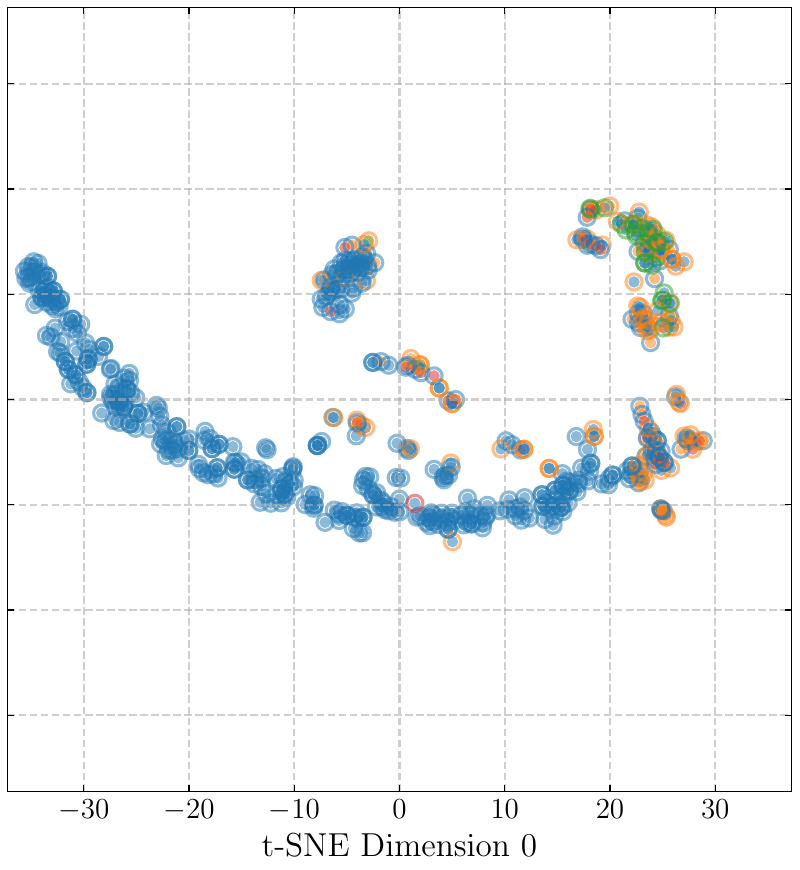}
    \includegraphics[height=0.45\textwidth]{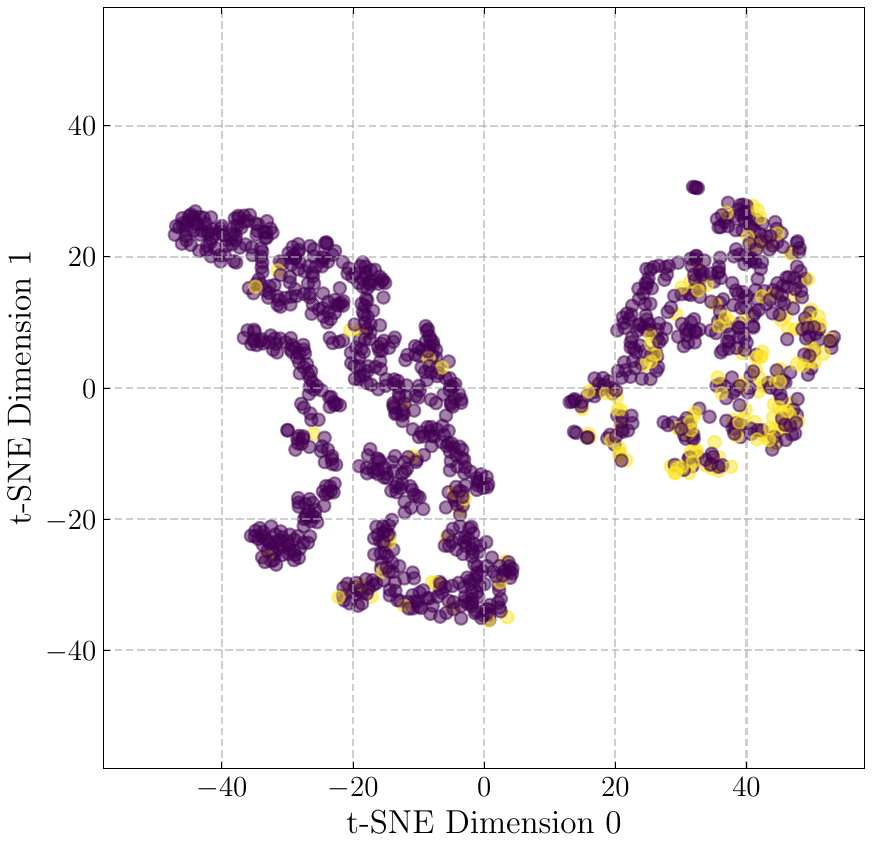}
    \includegraphics[height=0.45\textwidth]{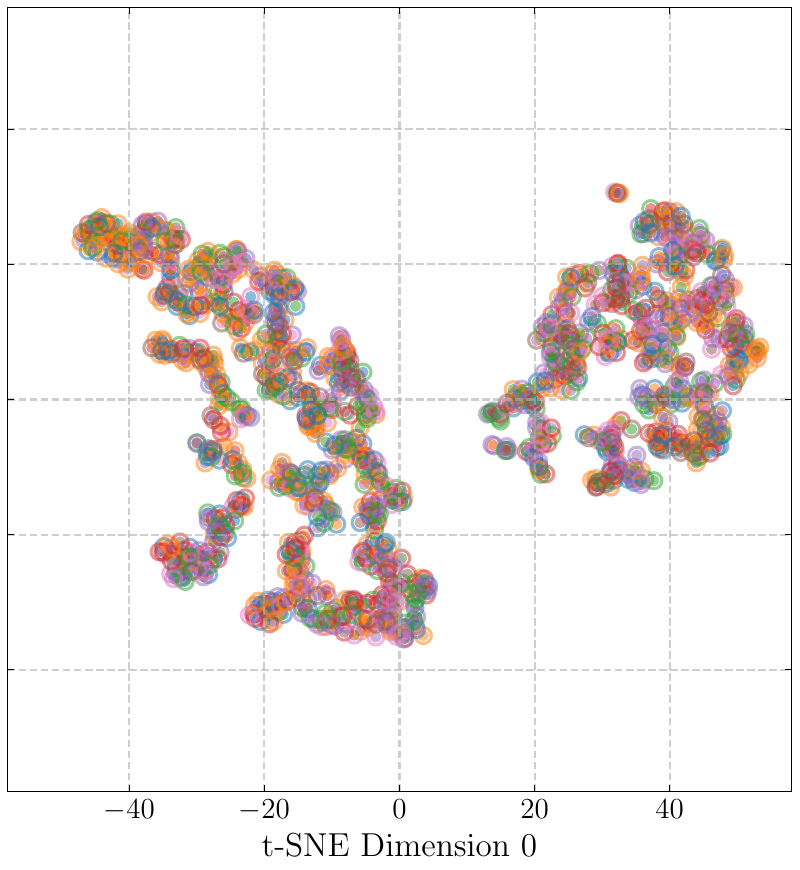}
    \includegraphics[height=0.45\textwidth]{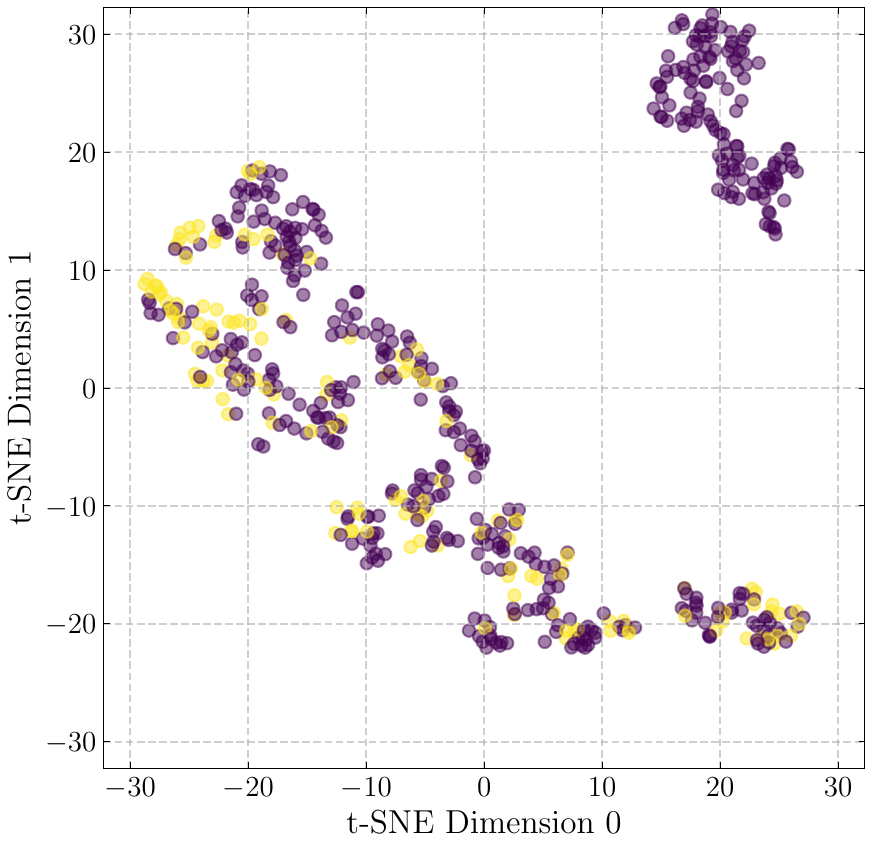}
    \includegraphics[height=0.45\textwidth]{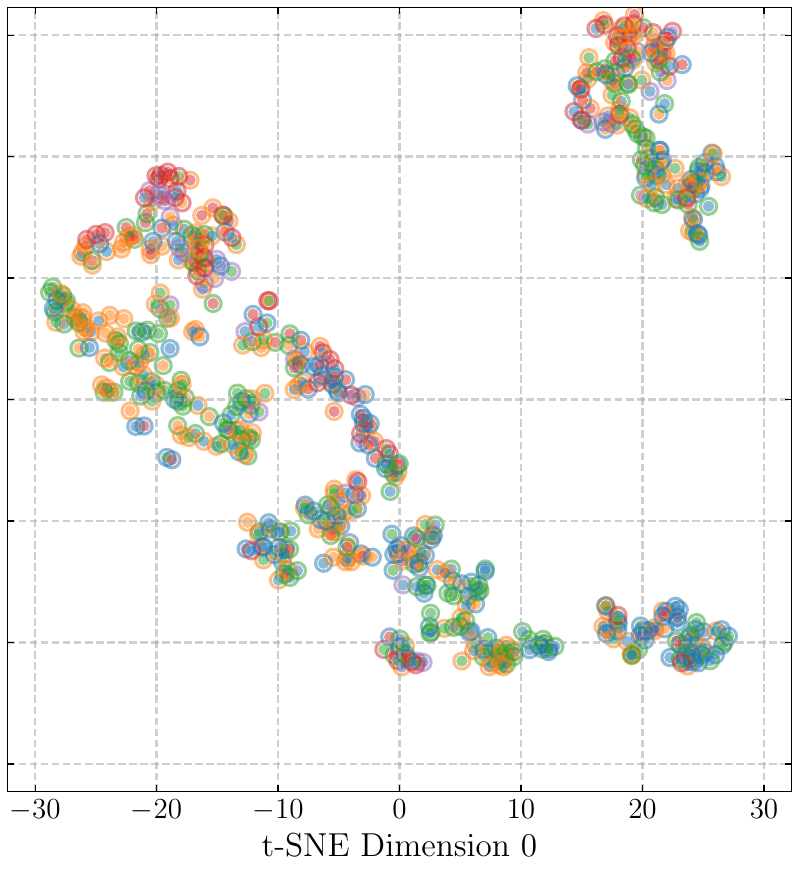}
    \caption{t-SNE plots \citep{maaten2008visualizing} of the edge features before the last triangle message passing layer for selected instance of the CP-Lib benchmark \cite{sorensen2024cplib}. In the plots on the left, each point is colored by whether the edge is cut (violet) or joined (yellow) in the same optimal solution. In the plots on the right, each point is colored based on the clusters of the corresponding nodes in an optimal solution. 
    The top row shows features for the instance ``cars'', the middle row for the instance ``ce50-40'' and the bottom row for the instance ``CPn35-3''.
    }
    \label{fig:tsne-layer2}
\end{figure}

\clearpage

\section{Additional Table}
\label{appendix:table}

\begin{table}[h]
\centering
\caption{Objective value and runtimes for solving instances of the CP-Lib benchmark \cite{sorensen2024cplib} with up to 200 nodes with our model. The column ``Optimal Value'' contains the best known objective value from literature. If it is not proven to be optimal, it is written in parentheses.}
\label{tab:approximate-solvers2}
\setlength{\tabcolsep}{12pt}
\begin{tabular}{l r r r}
\toprule
\textbf{Instance} & \textbf{Optimal Value} & \textbf{Our Value} & \textbf{Runtime [}s\textbf{]} \\
\midrule
corr40-1 & -2183 & -2183 & 0.502 \\
corr40-2 & -2206 & -2185 & 0.277 \\
corr40-3 & -2294 & -2284 & 0.237 \\
corr40-4 & -2544 & -2544 & 0.259 \\
corr40-5 & -2321 & -2311 & 0.253 \\
corr40-6 & -1749 & -1749 & 0.254 \\
corr40-7 & -2393 & -2393 & 0.248 \\
corr40-8 & -2271 & -2271 & 0.246 \\
corr40-9 & -2565 & -2565 & 0.234 \\
corr40-10 & -2161 & -2158 & 0.285 \\
corr60-1 & -3204 & -3200 & 0.852 \\
corr60-2 & -2836 & -2780 & 0.810 \\
corr60-3 & -4086 & -4086 & 0.808 \\
corr60-4 & -3530 & -3512 & 0.840 \\
corr60-5 & -4398 & -4372 & 0.815 \\
corr60-6 & -3617 & -3588 & 0.797 \\
corr60-7 & -4151 & -4141 & 0.793 \\
corr60-8 & -3951 & -3871 & 0.792 \\
corr60-9 & -3218 & -3136 & 0.801 \\
corr60-10 & -3305 & -3250 & 0.811 \\
corr80-1 & (-5026) & -4974 & 2.537 \\
corr80-2 & (-4534) & -4475 & 2.476 \\
corr80-3 & (-5003) & -4986 & 2.485 \\
corr80-4 & (-5236) & -5170 & 2.544 \\
corr80-5 & (-4443) & -4374 & 2.484 \\
corr80-6 & (-5154) & -5068 & 2.484 \\
corr80-7 & (-5389) & -5325 & 2.480 \\
corr80-8 & (-3838) & -3741 & 2.526 \\
corr80-9 & (-5336) & -5255 & 2.486 \\
corr80-10 & (-3847) & -3837 & 2.480 \\
\hline
\end{tabular}
\end{table}

\captionsetup{labelformat=empty}
\begin{table}[t]
\centering
\caption*{Table 4. (Continuation)}
\setlength{\tabcolsep}{12pt}
\begin{tabular}{l r r r}
\toprule
\textbf{Instance} & \textbf{Optimal Value} & \textbf{Our Value} & \textbf{Runtime [}s\textbf{]} \\
\midrule
bridges & -12585 & -12500 & 8.345 \\
cars & -185 & -185 & 0.178 \\
cetacea & -2757 & -2757 & 0.185 \\
companies & -3254 & -3254 & 20.636 \\
hayes-roth & -17524 & -16943 & 35.988 \\
lung-cancer & -837 & -837 & 0.171 \\
lymphography & -8696 & -8452 & 26.410 \\
micro & -1456 & -1456 & 0.236 \\
soybean-21 & -3562 & -3562 & 0.367 \\
soybean-35 & 0 & 0 & 0.379 \\
sponge & -5426 & -5426 & 2.039 \\
ta-evaluation & -20833 & -20831 & 28.495 \\
uno & -1449 & -1449 & 0.541 \\
uno\_1a & -19440 & -19413 & 34.381 \\
uno\_1b & -13030 & -13030 & 23.286 \\
uno\_2a & -45317 & -45187 & 34.222\\
uno\_2b & -20666 & -20666 & 24.369 \\
uno\_3a & -15499 & -15499 & 34.223 \\
uno\_3b & -1966 & -1966 & 25.725 \\
wildcats & -606 & -606 & 0.158 \\
workers & -383 & -383 & 0.190 \\
zoo & -1838 & -1838 & 6.625 \\
am-25-3 & -697400 & -697400 & 0.393 \\
am-25-10 & -1012225 & -1012225 & 0.341 \\
am-25-20 & -1461975 & -1461975 & 0.383 \\
am-50-3 & -10289800 & -10289800 & 5.924 \\
am-50-10 & -12861950 & -12861950 & 6.025 \\
am-50-20 & -16536450 & -16536450 & 6.562 \\
am-75-3 & -50574072 & -50574072 & 25.967\\
am-75-10 & -59314800 & -59314800 & 25.970 \\
am-75-20 & -71801552 & -71801552 & 25.966 \\
am-100-3 & -178198912 & -178198912 & 74.535 \\
am-100-10 & -207897888 & -207897888 & 74.374 \\
am-100-20 & -157409600 & -157409600 & 74.375 \\
CPn35-1 & -1094338 & -1093709 & 0.214 \\
CPn35-2 & -1244149 & -1242595 & 0.188 \\
CPn35-3 & -1254751 & -1252596 & 0.196 \\
CPn35-4 & -1193970 & -1192788 & 0.210 \\
CPn45-1 & -1892031 & -1888275 & 0.387 \\
CPn45-2 & -1841718 & -1840790 & 0.386 \\
CPn45-3 & -1931729 & -1929388 & 0.389 \\
CPn45-4 & -2231594 & -2229841 & 0.377 \\
CPn50-1 & (-2543544) & -2540516 & 0.513 \\
CPn50-2 & (-2184771) & -2182940 & 0.474 \\
CPn50-3 & (-2484518) & -2482190 & 0.441 \\
CPn50-4 & (-2293499) & -2289125 & 0.451 \\
\hline
\end{tabular}
\end{table}

\begin{table}[t]
\centering
\caption*{Table 4. (Continuation)}
\setlength{\tabcolsep}{12pt}
\begin{tabular}{l r r r}
\toprule
\textbf{Instance} & \textbf{Optimal Value} & \textbf{Our Value} & \textbf{Runtime [}s\textbf{]} \\
\midrule
CPn65-1 & (-3975105) & -3970775 & 1.568 \\
CPn65-2 & (-4016487) & -4012175 & 1.221 \\
CPn65-3 & (-3966284) & -3960474 & 1.314 \\
CPn65-4 & (-4114808) & -4109607 & 1.194 \\
CPn100-1 & (-9491009) & -9477322 & 6.472 \\
CPn100-2 & (-9569436) & -9555832 & 6.581 \\
CPn100-3 & (-9365902) & -9350392 & 6.443 \\
CPn100-4 & (-9317302) & -9303954 & 6.302 \\
rand100-5 & (-24449) & -1504 & 5.968 \\
rand100-100 & (-8744) & -30688 & 6.429 \\
rand200-5 & (-4590) & -3991 & 79.716 \\
rand200-100 & (-84667) & -76838 & 79.755 \\
boc\_1 & -296 & -296 & 0.341 \\
boc\_2 & -329 & -326 & 0.621 \\
boc\_3 & -356 & -356 & 0.316 \\
boc\_4 & -308 & -308 & 0.308 \\
boc\_5 & -338 & -338 & 0.330 \\
boc\_6 & -354 & -354 & 0.321 \\
boc\_7 & -334 & -334 & 0.337 \\
boc\_8 & -313 & -310 & 0.321 \\
boc\_9 & -331 & -331 & 0.347 \\
boc\_10 & -334 & -334 & 0.441 \\
boe\_91 & -474 & -473 & 0.555 \\
bur\_69 & -532 & -532 & 0.567 \\
bur\_73 & (-2762) & -2747 & 14.720 \\
bur\_75 & -503 & -502 & 0.993 \\
bur\_91 & -498 & -498 & 0.707 \\
can\_97 & -696 & -696 & 1.302 \\
cha\_86 & -532 & -532 & 0.573 \\
cha\_87 & -3507 & -3491 & 22.819 \\
gro\_80 & -287 & -285 & 0.256 \\
ira\_95 & -120 & -120 & 0.146 \\
kat\_97 & (-1108) & -1079 & 8.032 \\
kin\_80 & -259 & -259 & 0.191 \\
lee\_97 & -1011 & -1011 & 1.442 \\
mas\_97 & -167 & -167 & 0.175 \\
mcc\_72 & -257 & -257 & 0.205 \\
mil\_91 & -749 & -749 & 0.725 \\
nai\_96a & -791 & -791 & 1.387 \\
nai\_96b & -815 & -815 & 1.608 \\
nai\_96c & -769 & -769 & 1.344 \\
nai\_96d & -792 & -792 & 1.165 \\
rog\_05 & -818 & -816 & 1.102 \\
sei\_88 & -142 & -142 & 0.212 \\
sul\_91 & -124 & -124 & 0.191 \\
\hline
\end{tabular}
\end{table}

\begin{table}[t]
\centering
\caption*{Table 4. (Continuation)}
\setlength{\tabcolsep}{12pt}
\begin{tabular}{l r r r}
\toprule
\textbf{Instance} & \textbf{Optimal Value} & \textbf{Our Value} & \textbf{Runtime [}s\textbf{]} \\
\midrule
neg-c-00 & -1102 & -1097 & 0.538 \\
neg-c-10 & -1158 & -1152 & 0.535 \\
neg-c-20 & -1154 & -1147 & 0.456 \\
neg-c-30 & -1106 & -1106 & 0.422 \\
neg-c-40 & -949 & -943 & 0.527 \\
neg-c-50 & -851 & -850 & 0.521 \\
neg-c-60 & -683 & -682 & 0.443 \\
neg-c-70 & -548 & -548 & 0.437 \\
neg-c-80 & -425 & -424 & 0.524 \\
neg-s-80 & -576 & -565 & 0.885 \\
neg-tt-80 & (-728) & -719 & 1.750 \\
ce50-20 & -793 & -791 & 0.486 \\
ce50-30 & -570 & -568 & 0.471 \\
ce50-40 & -350 & -350 & 0.466 \\
ce50-50 & -164 & -164 & 0.430 \\
ce50-60 & -12 & -12 & 0.488 \\
ce60-20 & -1135 & -1131 & 0.744 \\
ce60-30 & (-808) & -801 & 0.802 \\
ce60-40 & -505 & -504 & 1.034 \\
ce60-50 & (-200) & -194 & 1.025 \\
ce60-60 & -19 & -19 & 0.996 \\
ce70-20 & -1542 & -1538 & 1.545 \\
ce70-30 & (-1095) & -1092 & 1.517 \\
ce70-40 & (-660) & -652 & 1.541 \\
ce70-50 & (-267) & -259 & 1.716 \\
ce70-60 & -8 & -8 & 1.518 \\
ce80-20 & -2003 & -1997 & 2.656 \\
ce80-30 & (-1421) & -1415 & 2.779 \\
ce80-40 & (-859) & -849 & 2.647 \\
ce80-50 & (-325) & -318 & 2.726 \\
ce80-60 & -25 & -25 & 2.724 \\
\hline
\end{tabular}
\end{table}

\end{document}